\documentclass[10pt,twocolumn,letterpaper]{article}

\usepackage{cvpr}
\usepackage{times}
\usepackage{epsfig}
\usepackage{graphicx}
\usepackage{amsmath}
\usepackage{amssymb}
\usepackage{graphicx}
\newcommand{\specialcell}[2][c]{%
  \begin{tabular}[#1]{@{}c@{}}#2\end{tabular}}

\usepackage{caption} 
\graphicspath{{figures/}}

\usepackage{authblk} 

\usepackage[breaklinks=true,bookmarks=false]{hyperref}

\cvprfinalcopy 

\def\cvprPaperID{6} 

\ifcvprfinal\fi
\begin{document}

\title{Okutama-Action: An Aerial View Video Dataset for Concurrent Human Action Detection\thanks{The work was conducted while the first 3 authors were participating in the NII International Internship Program, and the 4th was an Invited Researcher at NII.}}

\author{
Mohammadamin Barekatain$^{1}$, Miquel Mart\'{i}$^{2,3}$, Hsueh-Fu Shih$^{4}$, Samuel Murray$^{2}$, Kotaro Nakayama$^{5}$, Yutaka Matsuo$^{5}$, Helmut Prendinger$^{6}$\\
$^1$Technical University of Munich, Munich, $^2$KTH Royal Institute of Technology, Stockholm, $^3$Polytechnic University of Catalonia, Barcelona, $^4$National Taiwan University, Taipei, $^5$University of Tokyo, Tokyo, $^6$National Institute of Informatics, Tokyo\\
{\tt\small m.barekatain@tum.de, miquelmr@kth.se, r03945026@ntu.edu.tw, samuelmu@kth.se, nakayama@weblab.t.u-tokyo.ac.jp, matsuo@weblab.t.u-tokyo.ac.jp, helmut@nii.ac.jp}
}

\maketitle
\ifcvprfinal\thispagestyle{empty}\fi

\begin{abstract}
    Despite significant progress in the development of human action detection datasets and algorithms, no current dataset is representative of real-world aerial view scenarios. We present Okutama-Action, a new video dataset for aerial view concurrent human action detection. It consists of 43 minute-long fully-annotated sequences with 12 action classes. Okutama-Action features many challenges missing in current datasets, including dynamic transition of actions, significant changes in scale and aspect ratio, abrupt camera movement, as well as multi-labeled actors. As a result, our dataset is more challenging than existing ones, and will help push the field forward to enable real-world applications.
\end{abstract}

\section{Introduction}
\label{sec:introduction}
With the increased use of unmanned aerial vehicles (UAVs) for tasks such as surveillance, delivery and search and rescue, we believe that a better understanding of human actions from an aerial view is important. For example, in surveillance tasks, it can be essential to recognise actions and track the actors in order to detect anomalies. Likewise, for search and rescue missions, being able to distinguish a person's action could help the system understand if that person is in need of help. Although several new action recognition datasets have been presented over the last years, we argue that none is suitable for being used with UAVs. Not only is the view angle and scale of objects different from UAVs cameras, but available datasets also suffer from not being representative of common outdoor actions. 

\begin{figure*}[htb]
     \centering
     \includegraphics[width=0.25\textwidth]{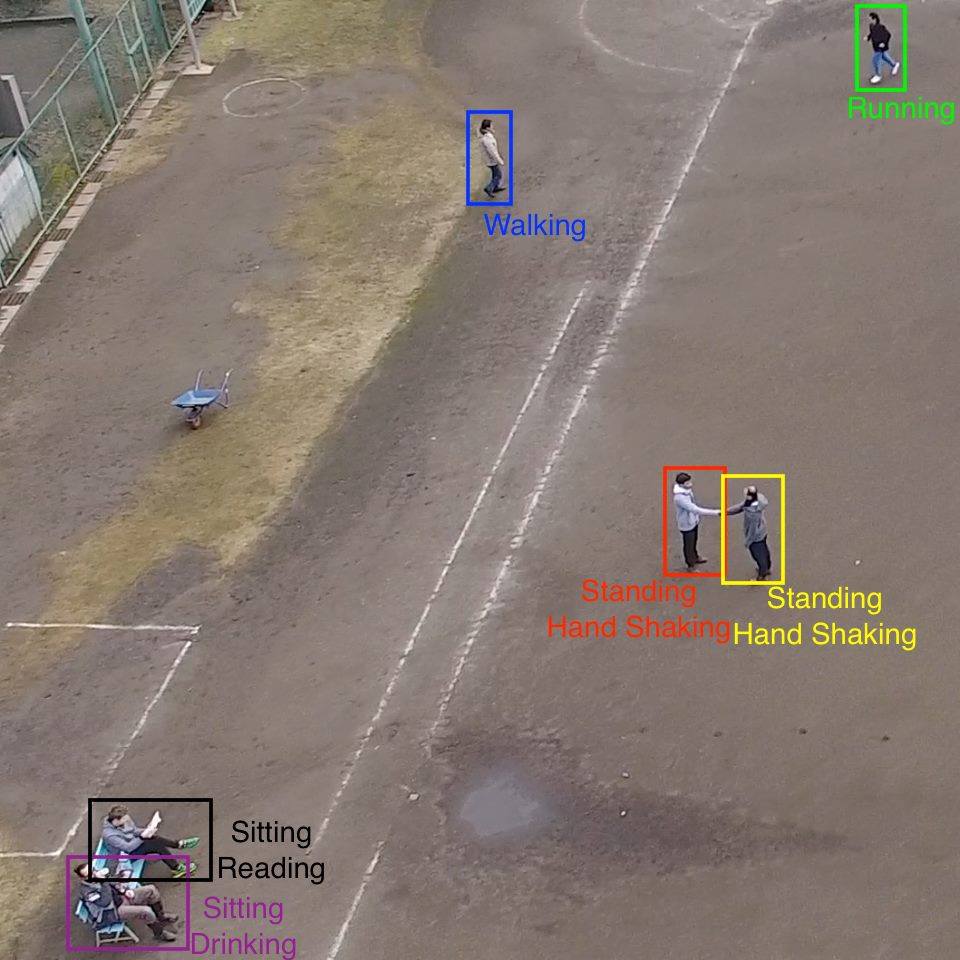}
     \qquad
     \includegraphics[width=0.25\textwidth]{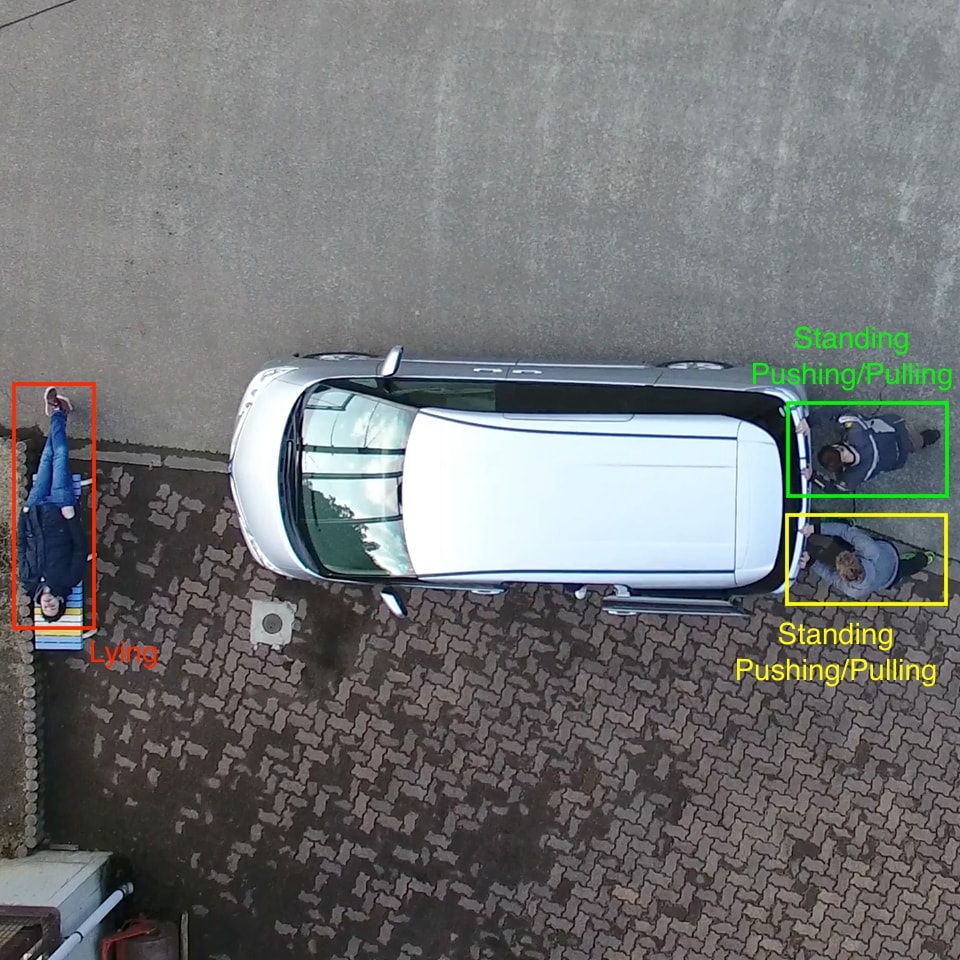}
     \qquad
     \includegraphics[width=0.25\textwidth]{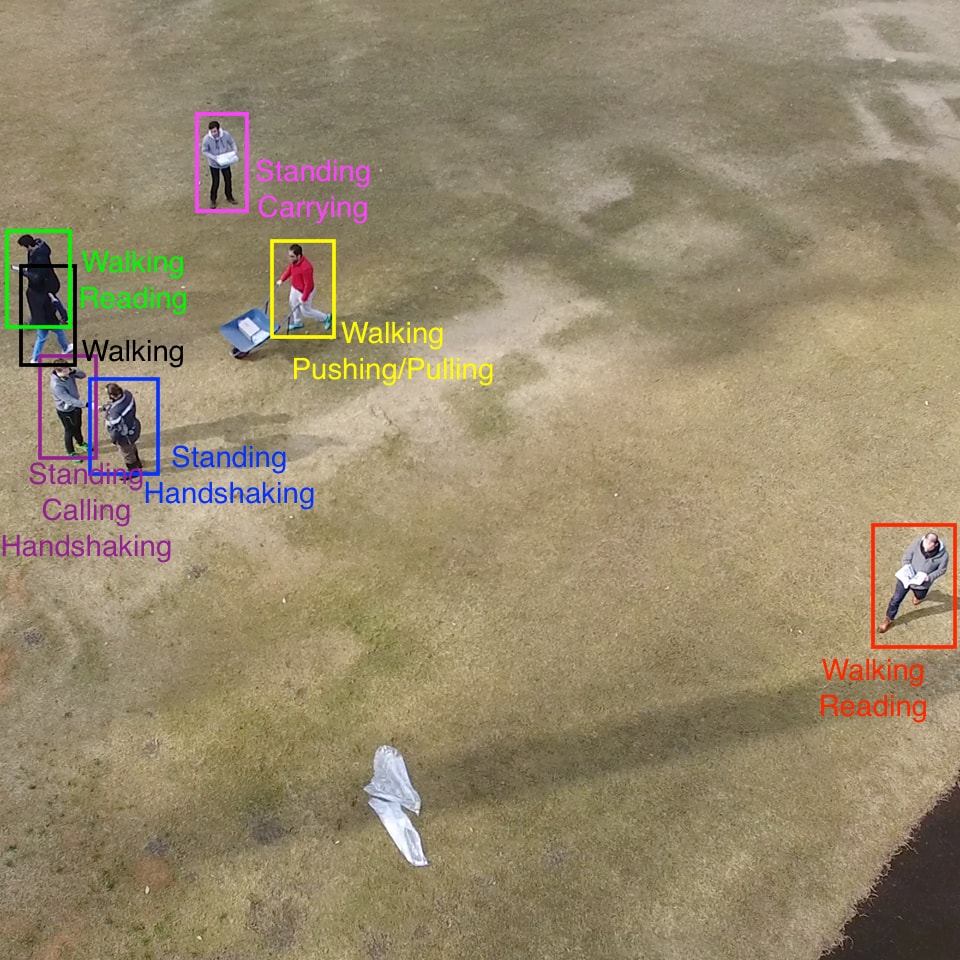}
     \qquad
     \includegraphics[width=0.25\textwidth]{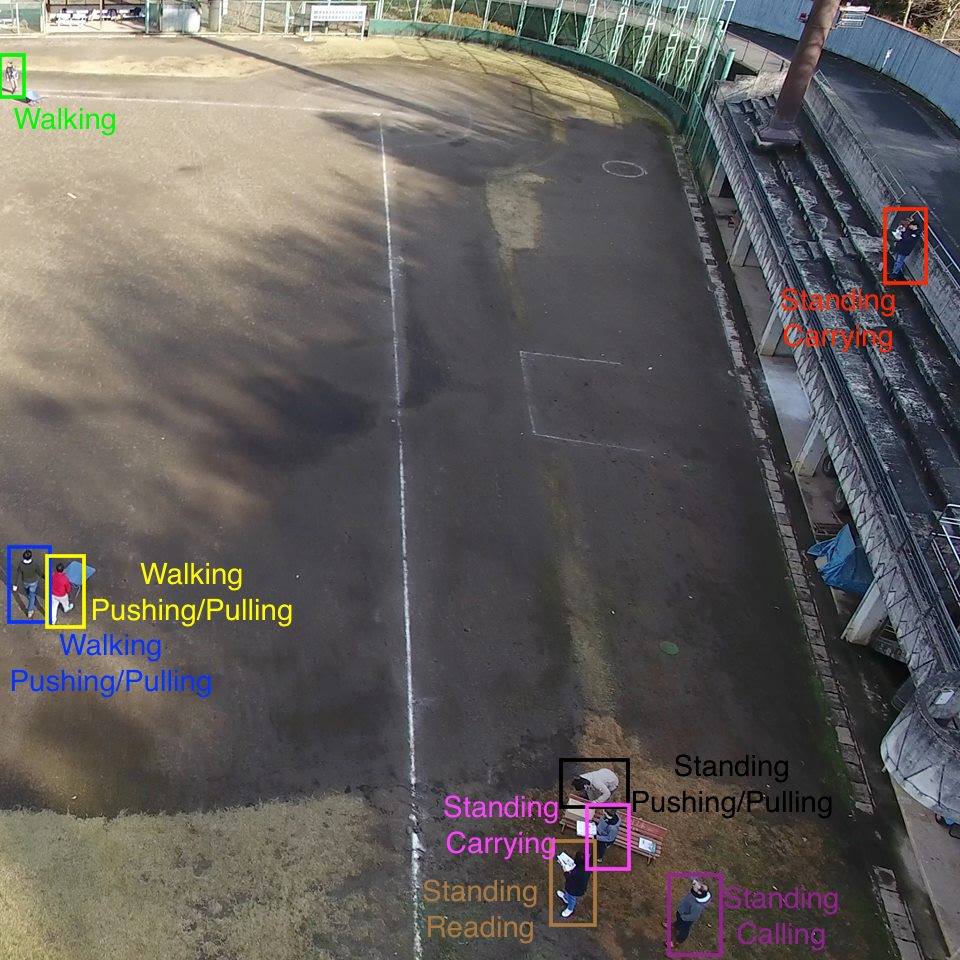}
     \qquad
     \includegraphics[width=0.25\textwidth]{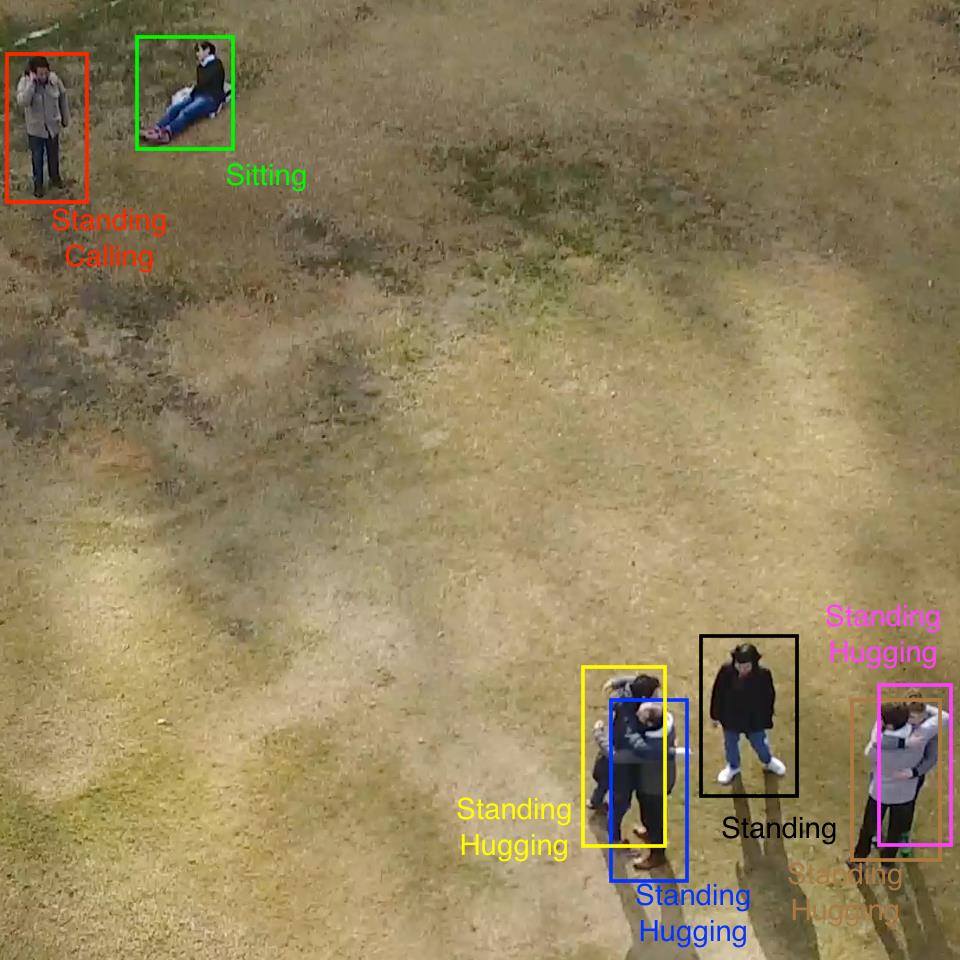}
     \qquad
     \includegraphics[width=0.25\textwidth]{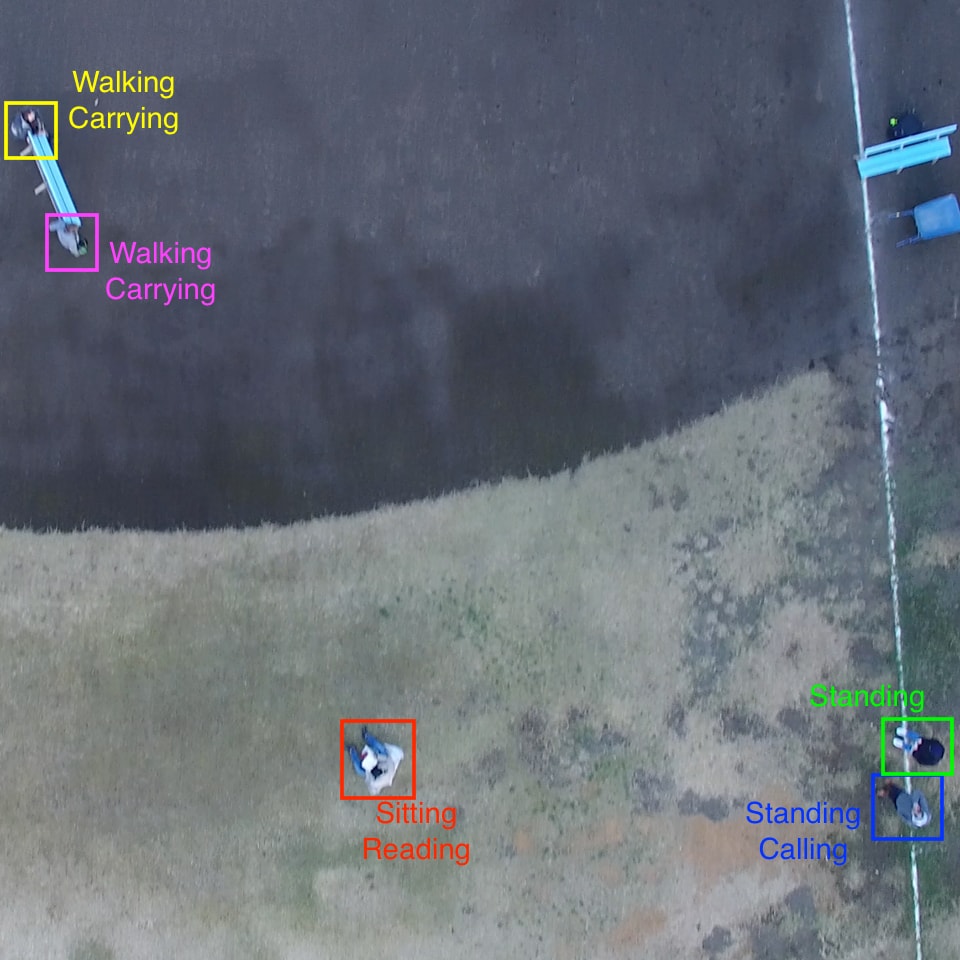}
     \caption{Sample frames of Okutama-Action dataset. These are cropped versions of original frames.}
     \label{fig:Sample_frames}
\end{figure*}

To address this, we present a new dataset, Okutama-Action, captured from UAVs flying at different altitudes and at different angles, to get a diverse set of sequences. Each sequence is much longer than those in other datasets, which makes them more similar to real world tasks where objects must be tracked over extensive time periods. In total, 12 action classes are used, deemed to be typical outdoor actions. Since basic actions like \emph{sitting} and \emph{walking} are annotated, all humans are labeled in each frame and they may have more than one labeled action. Compared to previous datasets, our action classes are more difficult to tell apart visually, because there are less distinguishing features exterior to the actor, such as change in environment. This is also shown by training and evaluating a state-of-the-art action detection model, which performs worse on Okutama-Action than on other datasets. This indicates that better action detection models have to be developed for use in real-world applications. Figure \ref{fig:Sample_frames} illustrates some examples of our dataset.

The outline of this paper is as follows: first, a review of currently available datasets for spatio-temporal human action detection is presented. Second, the details and the design choices behind our Okutama-Action dataset are presented, with comparisons to the reviewed datasets. Last, an action detection model is trained and evaluated on our dataset, and its performance is compared to that on other datasets.

\section{Related work}
\label{sec:related_work}

Reviews of current datasets for human action detection \cite{action-survey,survey2013} indicate a lack of aerial video sequences captured from a mobile platform. The same reviews also ascertain the lack of several challenges existing in real-world airborne scenarios, including dynamic transition of actions, significant changes in scale and aspect ratio and abrupt camera motion. Furthermore, currently available datasets are limited in at least one of these aspects: action types, number of concurrent actors, temporal length of actions, diversity of concurrent actions, video resolution and sequence duration.

We briefly review some of the relevant action detection datasets which include spatial annotation of actions. The UCF Sports dataset \cite{ucf-sports-1,ucf-sports-2} contains 150 sports videos for 10 action categories and the J-HMDB dataset \cite{Jhmdb} consists of 928 videos with 21 actions. Although these datasets include certain challenges such as camera motion, it is still relatively easy to recognize the actions by observing only the scene or the pose of the actor in a single frame \cite{vu2014predicting}. In addition, videos are of low resolution, contain only a single action and are trimmed to the action's duration, which is generally short. 

The UCF-101 dataset\footnote{The original UCF-101 dataset, that contains over 10K videos for 101 actions does not include spatio-temporal annotations. This annotation was later provided in THUMOS’13 challenge \cite{thumos13} for a subset of UCF101.} \cite{ucf101} with 24 action classes and 3207 sequences, is the largest and most diverse dataset to date which includes challenges such as camera motion, and changes in scale and viewpoint. Unfortunately, the actions covered in this dataset are not typical in aerial scenarios as it mainly contains indoor and sports actions. Additionally, although video sequences may contain concurrent actors, they all perform the same action. Furthermore, videos are of low resolution (320x240), have short duration, and in most of them, the actions lasts more than 80\% of the video duration \cite{DALY}.

Two of the few datasets which includes video samples of concurrent actions from different categories are LIRIS-HARL \cite{LIRIS} and DALY \cite{DALY} dataset, both with 10 action classes. LIRIS-HARL is captured in an office environment with different cameras, including a moving camera mounted on a mobile robot; videos in DALY are from YouTube. Unfortunately, in both datasets action categories are not common to aerial scenarios and there is no dynamic transition of actions, i.e.\ each actor performs a single action in each video clip. Moreover, in LIRIS-HARL, videos have low resolution (720x576) and short duration, and DALY is weakly-annotated, meaning that for each temporal action instance, only 5 uniformly sampled frames are annotated.  

Another work worth mentioning is UT-Interaction dataset \cite{UT-in}, which consists of 20 video sequences of continuous executions of 6 action classes. Actions from different categories may occur concurrently, there is dynamic transition of actions and videos are not trimmed to action duration. Unfortunately, the total number of videos and frames is low, which makes the dataset inappropriate for deep learning models, and various real-world challenges are missing. For example, action types are limited to interaction between two humans, the camera is static, and there is no partial occlusion of actions.

In recent years, various datasets for human action recognition have been constructed, such as ActivityNet \cite{activitynet}, Sports-1M \cite{Sports-1M}, HMDB51 \cite{hmdb}, MPII Cooking \cite{cooking-1,cooking-2}, Olympic Sports \cite{Olympic-sports}, Hollywood \cite{Hollywood2,Hollywood1} and MERL Shopping \cite{merl}. However, the annotation of these datasets does not include spatial location of actions. Moreover, instead of localization of actions for each person, some video benchmarks \cite{virat,shu2015} are intended for assessing the performance of event recognition algorithms, where an event can be composed of multiple complex human activities.  Also related to our work is Stanford drone dataset \cite{stanford} intended for trajectory forecasting and multi-target tracking in aerial view videos.

Okutama-Action provides 43 fully-annotated sequences which are of use to train and evaluate models for spatio-temporal detection of multiple concurrent human actions from different categories in the video footage of a mobile aerial platform. Our dataset is unique in the following aspects: an aerial view dataset that contains representative samples of actions in real-world airborne scenarios; dynamic transition of actions where, in each video, up to 9 actors sequentially perform a diverse set of actions; and a real-world challenge of multi-labeled actors where an actor performs more than one action at the same time. Additionally, our dataset has a significant increase compared to previous datasets, in number of actors and concurrent actions (up to 10 actions/actors), as well as video resolution (3840x2160) and sequence length (one minute on average).

\section{Okutama-Action development}
\label{sec:dataset}
All videos of Okutama-Action are captured from UAVs (DJI Phantom 4) at a baseball field in Okutama, Japan. In this section, we describe the selected action categories as well as the data collection settings. We then explain the annotation process and summarize the properties of our dataset. See Figure \ref{fig:Sample_frames} for example frames of our dataset.

\subsection{Dataset design and collection}
\begin{figure}[hb]
    \centering
    \includegraphics[width=.47\textwidth]{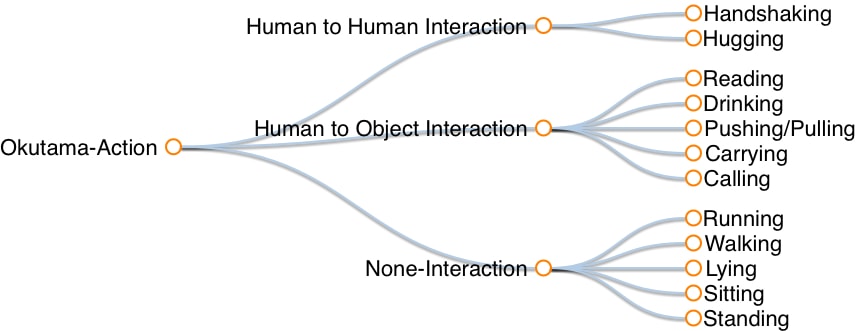}
    \caption{Categorization of action classes in Okutama-Action dataset}
    \label{fig:taxonomy}
\end{figure}

\textbf{Action selection} In order to collect video samples of human actions that are representative of everyday outdoor actions, we analyzed the video footage of low-altitude UAVs which led to a selection of 12 actions, including \textit{Reading}, \textit{Handshaking}, \textit{Drinking} and \textit{Carrying}. Inspired by \cite{ucf101}, we group these actions into 3 types: 1) Human to human interaction; 2) Human to object interaction; 3) None-interaction. Figure \ref{fig:taxonomy} displays all action classes and their corresponding groups.

\textbf{UAVs configuration} We experimented with different altitudes and camera angles for capturing videos of these actions, in order to find the proper settings for our UAVs during the data collection, ensuring that the actions are clearly visible and distinguishable. Based on this experiment, we decide the altitude range to be from 10 to 45 meter and camera angle to be either 45 or 90 degree.

\textbf{Data collection} In order to ensure that the sequences of our dataset are representative of real-world aerial applications, scripts for 22 scenarios were written in which up to 9 actors participate. We attempt to include various transitions of actions that can happen in real-world, while also having variability in execution style and in number of actors. For example, actors carry different items and read books of different sizes, and there exists crowded frames with 9 actors as well as deserted frames with no actor. Furthermore, in some scenarios, the actors were asked to perform random actions of their choice in order to increase the diversity.

In addition, a separate set of scenarios were written for our 2 UAV pilots in order to make sure we have variety in viewpoint. For example, in some sequences the UAV is still and only spinning while in others it may be moving with a top-down camera angle. The dataset also includes some metadata for each video sequence, namely camera angle, speed and altitude. Furthermore, it was our goal to include common challenges existing in the video footage of airborne platforms such as partial occlusion of actors as well as changes in scale, aspect ratio and camera speed.

Each scenario, with the exception of one, was captured using 2 UAVs (of different configuration) at the same time, which, together with the metadata, are of use for action detection algorithms comparison.
Data collection was done in two different lighting conditions (sunny and cloudy) at a baseball field and the actors were a group of researcher at our lab. Video sequences were recorded at 4K resolution and 30 FPS using a high performance camera mounted on an adjustable, integrated gimbal system on the UAVs. 


\subsection{Dataset annotations}
\label{subsec:datasetannotations}
We use VATIC \cite{VATIC}, an open source video annotation tool, integrated with Amazon Mechanical Turk to manually annotate the videos at 10 FPS. The annotations are then linearly interpolated to 30 FPS. The bounding boxes and their corresponding action labels are reviewed and adjusted by members of our group.

In the original annotation, bounding boxes may have more than one label, since an actor may naturally perform multiple actions at the same time. However, current action detection algorithms are limited to detect a single action; hence, we also provide an annotation set in which the bounding boxes have a single label. To do so, we first ranked the action categories based on their group type and number of instances that they have in our dataset (the lower the number of instances is, the higher its rank). The None-interaction actions have the lowest rank since by default an actor is always performing an action of this type (e.g. \textit{Reading} while \textit{Sitting}). Finally, for each bounding box, we keep the action with the highest rank.

\subsection{Dataset summary and statistics}
\label{subsec:datasetsummary}
\begin{figure*}[ht]
    \centering
    \includegraphics[width=0.46\textwidth]{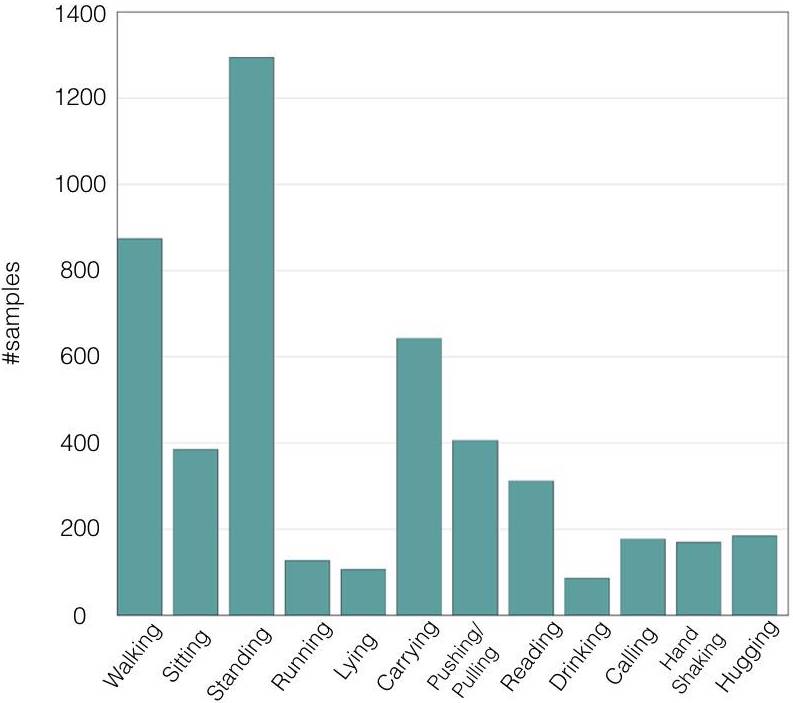}
    \includegraphics[width=0.48\textwidth]{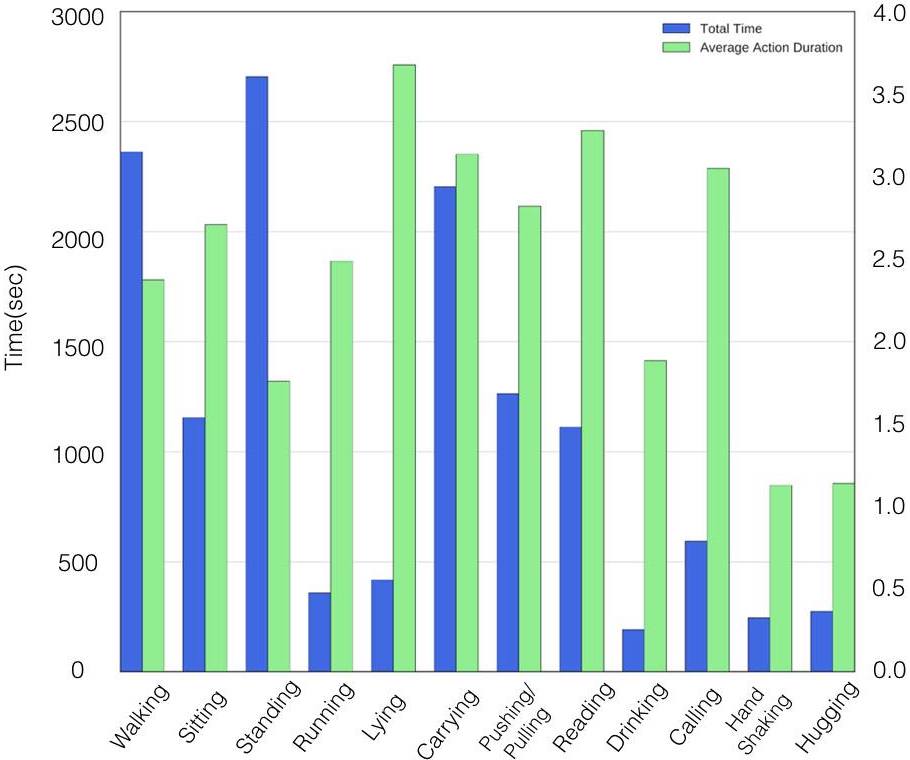}
    \caption{Okutama-Action statistics. Left: Number of samples (instances) per action class. Right: Total duration of actions for each action class is illustrated using the blue bars. The average action duration for each class is depicted in green.}
    \label{fig:stats}
\end{figure*}

Our new Okutama-Action dataset contains a total of 43 video sequences at 30 FPS and 77365 frames in 4K resolution. These sequences were recorded using 2 UAVs flying at altitudes varying between 10-45 meters and with camera angle of 45 or 90 degrees. Figure \ref{fig:stats}, on the left side, demonstrates the number of samples per action class, and the chart shown on the right side illustrates total duration of actions (blue) and the average action duration (green) for each action category. These statistics are calculated from the single-label annotation of the dataset.

\textbf{Dataset split for training and testing set} For evaluation purposes, we split our dataset into two distinct sets: 1) train-val set, consisting of 33 video sequences; 2) test set, consisting of 10 video sequences. This split is designed in a way to make sure that the two sequences from the same scenario are in the same set. If this were not the case, the test data would not be completely unseen - for example, a model could memorize the relative positions of actors, and the specific action transitions for a given sequence. Moreover, the split ensures that diverse challenges exists in the test set, which is important for evaluating the robustness of action detection models. For example, not all sequences have a change in camera angle or altitude, but the test set should properly assess a model's performance under these circumstances. Lastly, since the configuration of the UAVs (speed, altitude and camera angle) differs between sequences, it is possible to identify the UAV configuration that gives the best performance for a given model, so that during deployment in real-world applications the optimal configuration can be used.

\textbf{Comparison with other datasets} Some features of our dataset is compared to existing datasets in Table \ref{tab:comparison}. Okutama-Action is the second largest fully-annotated dataset for concurrent human action detection task after UCF101. We would like to emphasise that no other action detection dataset includes samples which are representative of real-world airborne scenarios. To the best of our knowledge, Okutama-Action is the first dataset for spatio-temporal action detection that includes multi-labeled actors. Moreover, our dataset has made significant progress in terms of number and diversity of concurrent actions as well as video resolution and sequence duration.

\begin{table*}
    \centering
    \begin{tabular}{|c|c|c|c|c|c|c|c|c|} 
        \hline
        Dataset & Year & \specialcell{Number of\\Actions} & \specialcell{Total\\Frames} & \specialcell{Average\\Video Dur.} & Resolution & \specialcell{Concurrent\\Actions}  & Resource \\\hline
        UCF Sports \cite{ucf-sports-1,ucf-sports-2} & 2008 & 10 & 10K & 5.8s & 690x450 & No & TV, Movies \\
        UT-Interaction \cite{UT-in} & 2010 & 6 & 36K & \textbf{60s} & 720x480 & \textbf{Yes} & Actor Staged \\
        UCF-101 \cite{ucf101} & 2012 & \textbf{24} & \textbf{558K} & 5.8s & 320x240 & \textbf{Yes} & YouTube \\
        J-HMDB \cite{Jhmdb} & 2013 & 21 & 32K & 1.4s & 320x240 & No & Movies, YouTube \\
        LIRIS-HARL \cite{LIRIS} & 2014 & 10 & 64K & 15.2s & 720x576 & \textbf{Yes} & Actor Staged \\
        Okutama-Action & 2017 & 12 & 77K & \textbf{60s} & \textbf{3840x2160} & \textbf{Yes} & Actor Staged \\ \hline
    \end{tabular}
    \caption{Comparison of Okutama-Action dataset with current fully-annotated spatio-temporal human action detection datasets.}
    \label{tab:comparison}
\end{table*}

\section{Experimental results}
\label{sec:results}
In this section, we show how to adapt a simple model designed for object detection to the tasks of action detection, and evaluate it on both tasks with our Okutama-Action dataset. First, we describe the Single Shot MultiBox Detector (SSD) \cite{liu2016ssd}, we explain how we use it for detecting pedestrians in our dataset and show the results we get. Second, we explain how we use the same model for action detection and show our results. The results reported in this section were computed using the split detailed in Section \ref{subsec:datasetsummary}. All experiments were carried out using the SSD Caffe fork with CUDA 7 and two Nvidia K40 GPUs.

\subsection{Pedestrian detection}
\textbf{Model description} SSD is a unified object detector implemented in a single network. Having inference speed in mind, the space of possible boxes is discretized coarsely into a set of default boxes and then their coordinates are refined to tightly surround the object. This is done with different aspect ratios and at different scales by taking features from different stages of the network. For each (refined) default box, the network predicts the likelihood of having an object of a given class in it. As other single-shot approaches, e.g. YOLO \cite{redmon2016you}, SSD avoids having a separate region proposal generation step and gives a prediction with a single forward pass of the network.

\textbf{Training strategy} We train the SSD model based on VGGnet \cite{Simonyan14c} with an image input size of 512x512 pixels, following the original strategy of 20000 iterations with a learning rate of $10^{-4}$ and the rest of hyperparameters set to the default values of SSD. In order to use our dataset for the pedestrian detection task, we give all bounding boxes the label \textit{Pedestrian}. This way we validate our choice‚ as we expect this to be an easier task; we believe that if good performance on pedestrian detection is not achieved, the model can not successfully be used for action detection. The resulting detections could be used by a Multiple Object Tracking algorithm directly.

\begin{figure}[htb]
     \centering
     \includegraphics[width=0.45\textwidth]{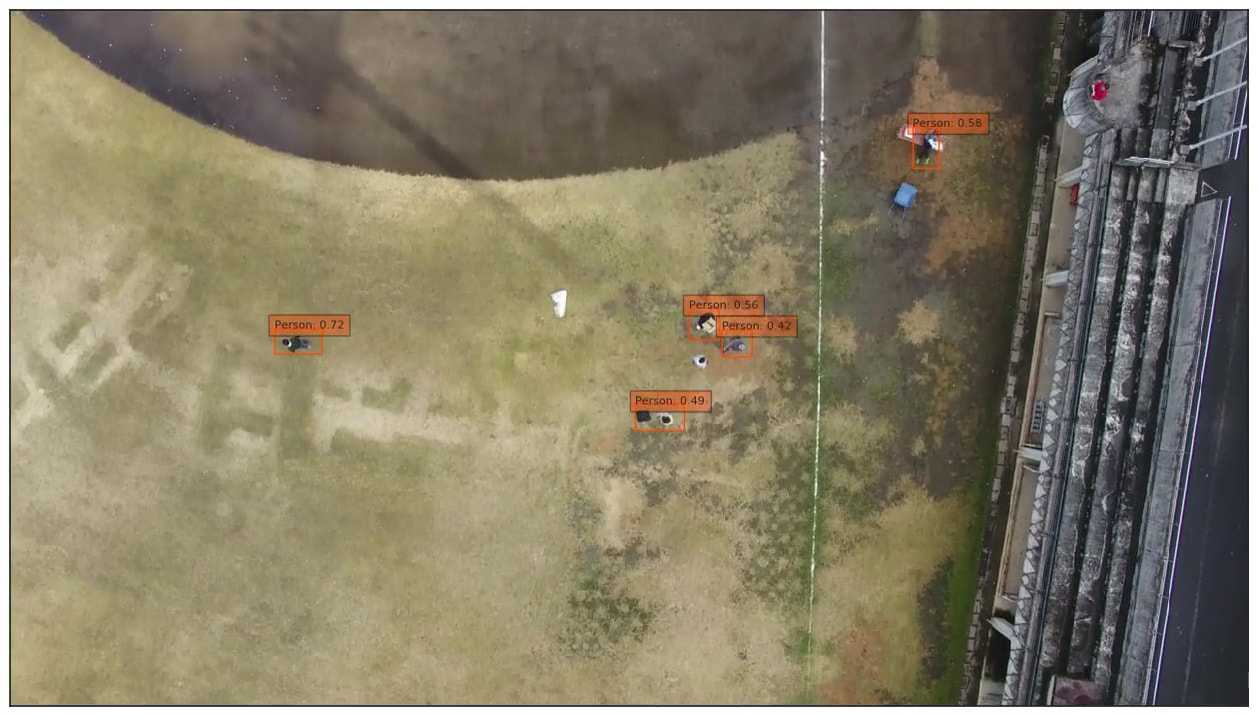}
     \qquad
     \includegraphics[width=0.45\textwidth]{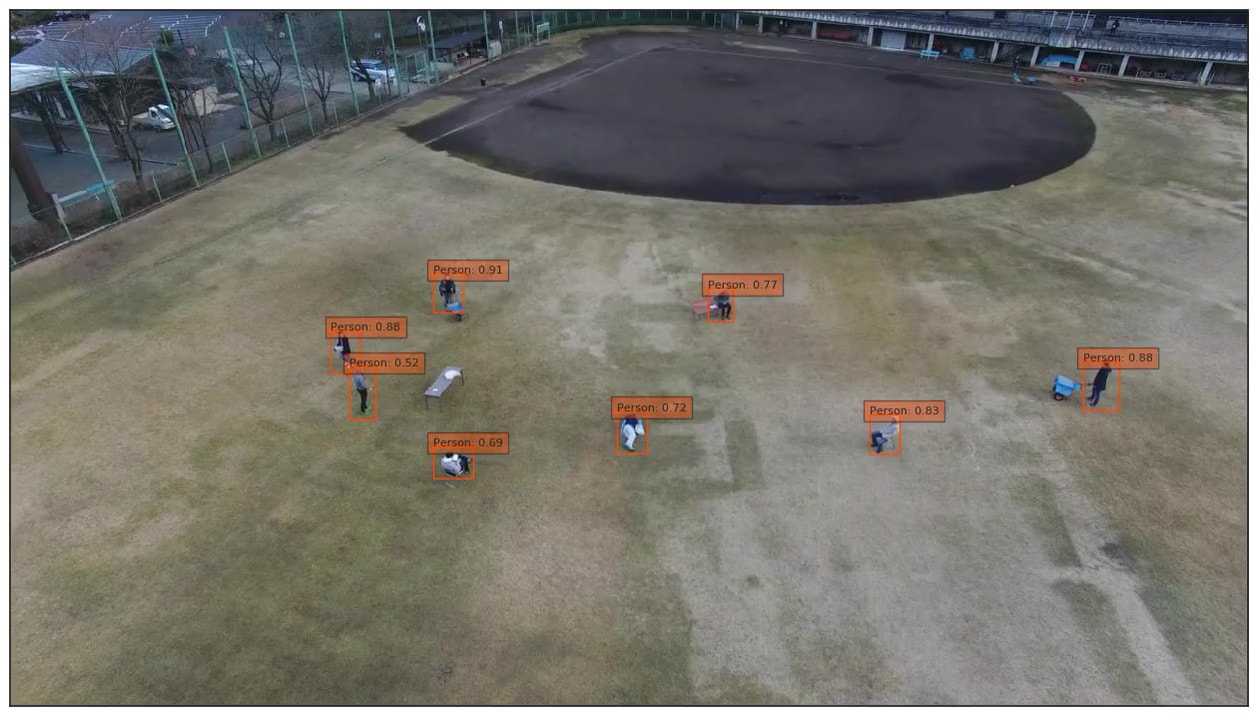}
     \caption{Sample detections of SSD model for pedestrian detection. Detections with confidence score higher than 0.4 are shown. Frames from Okutama-Action test set.}
     \label{fig:peddetsample}
\end{figure}

\textbf{Results} We use mean Average Precision at 0.5 IoU threshold (mAP@0.5) as an evaluation metric as is commonplace in object detection tasks \cite{redmon2016you,liu2016ssd,Huang2016SpeedaccuracyTF}. In our case, with only one class, it is simply the Average Precision for the class \textit{Pedestrian} and  we get a value of $72.3\%$ on the test set of Okutama-Action dataset. Figure \ref{fig:peddetsample} shows two selected frames with the detection results of the best model. We note that as reported in \cite{Huang2016SpeedaccuracyTF}, the model performs poorly when pedestrians are too small, which we determine happens when the altitude of the UAVs is higher than approximately 30 meters. 

\subsection{Action detection}

\textbf{Model description} The action detection model we use follows \cite{singh2016online} to get spatio-temporal action localisation and prediction. This model follows a two-stream approach which can be divided in three steps: SSD is the object detector of choice used in the first step of \cite{singh2016online} to get the location and class of the actions as detection boxes, in both natural RGB and optical flow streams, for each frame. The second step merges the detections and classification scores of both streams to combine the appearance and motion cues from the natural and optical flow images. In the third step, the sequences of detections are used to incrementally construct action tubes - sequences of detections pertaining to a single action - giving temporal and spatial consistency of the actions across frames and delimiting their duration in an online fashion. Here, for simplicity, we limit our evaluation to the first step. We use the annotation set with only one concurrent action, described in Section \ref{subsec:datasetannotations}, as SSD cannot handle multiple labels.

\textbf{Training strategy} We train the model with different input image sizes, as we expect a larger size to result in increased accuracy, though at the cost of longer training and inference times. When setting the input image size to 512x512 we train for 34000 iterations with an initial learning rate of $10^{-3}$, which we divide by 10 after 19000 and 30000 iterations. For the larger input image size of 960x540, which has the same aspect ratio as the original video, we train the model for 60000 iterations and use an initial learning rate of $10^{-4}$, which is divided by 10 after 40000 iterations. We also train a model on the optical flow images of size 512x512, with the same parameters as the RGB 512x512 model.

\begin{figure}[htb]
     \centering
     \includegraphics[width=0.45\textwidth]{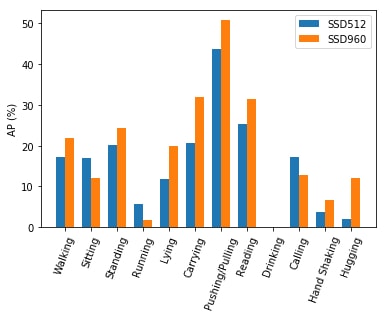}
     \caption{Per-class Average Precision for the models trained for action detection evaluated on Okutama-Action test set. Blue for the model with input size 512x512; Orange for the model with input size 960x540.}
     \label{fig:perclassres}
\end{figure}

\begin{figure*}[htb]
     \centering
     \includegraphics[width=0.45\textwidth]{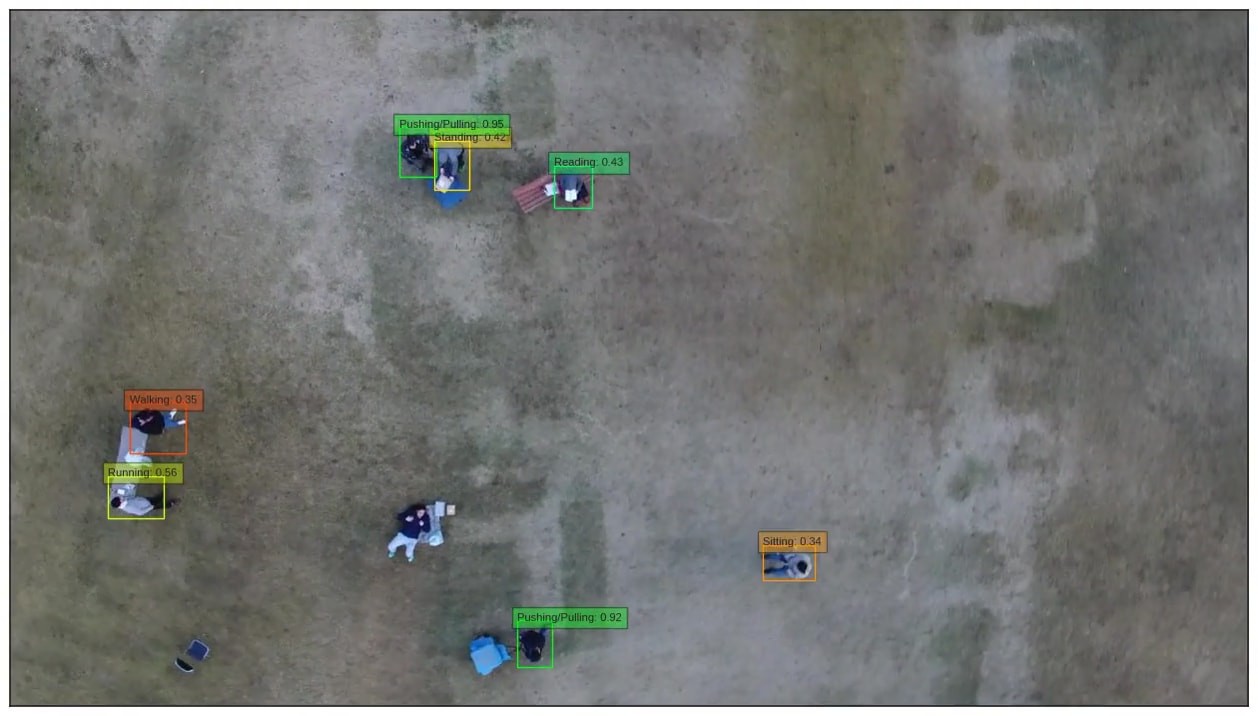}
     \qquad
     \includegraphics[width=0.45\textwidth]{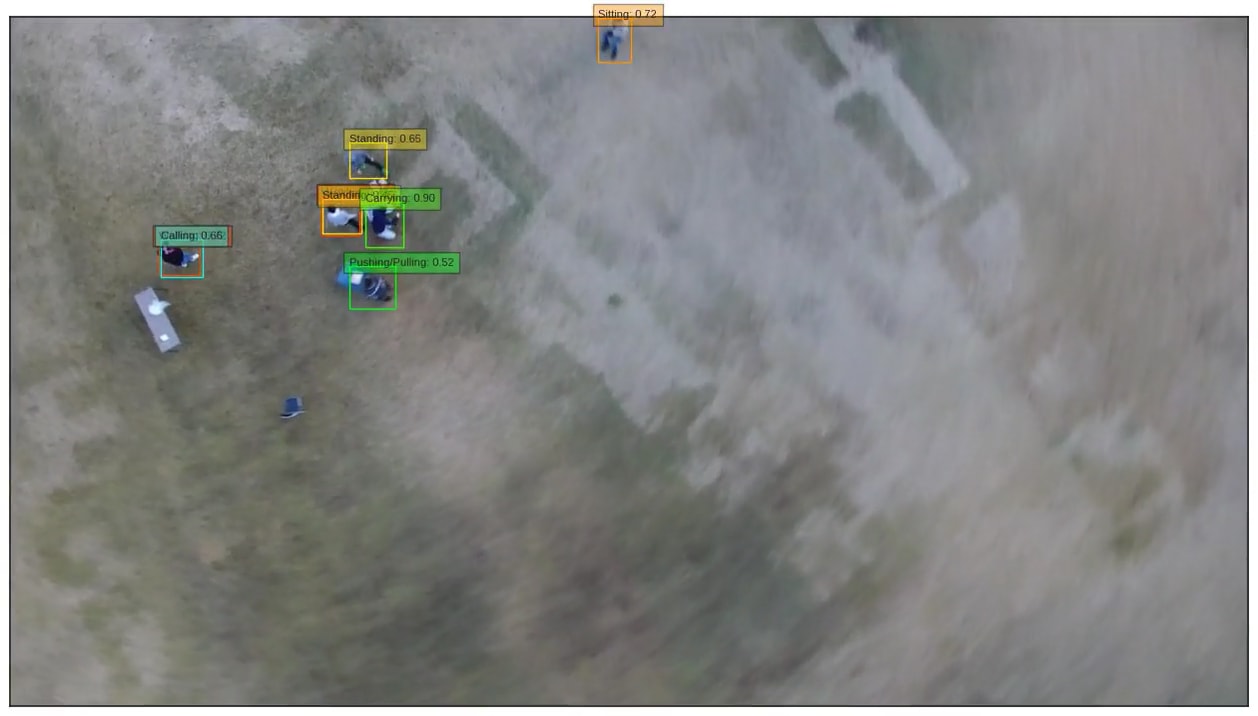}
     \qquad
     \includegraphics[width=0.45\textwidth]{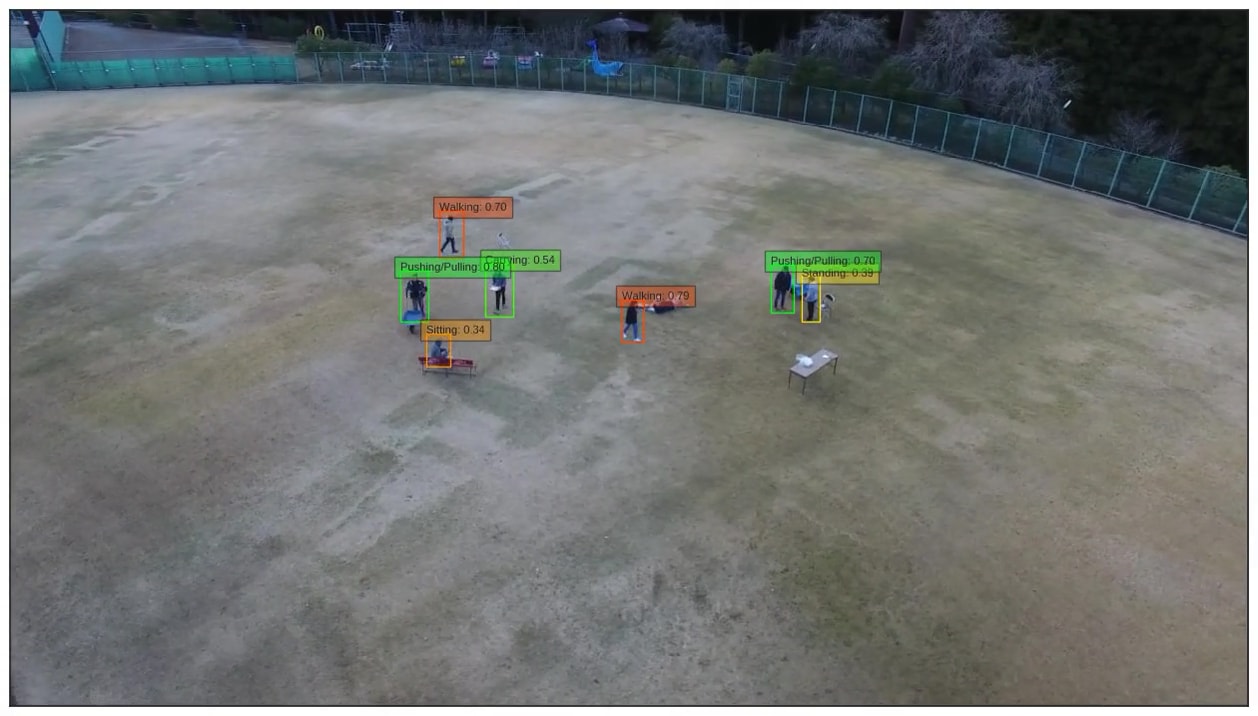}
     \qquad
     \includegraphics[width=0.45\textwidth]{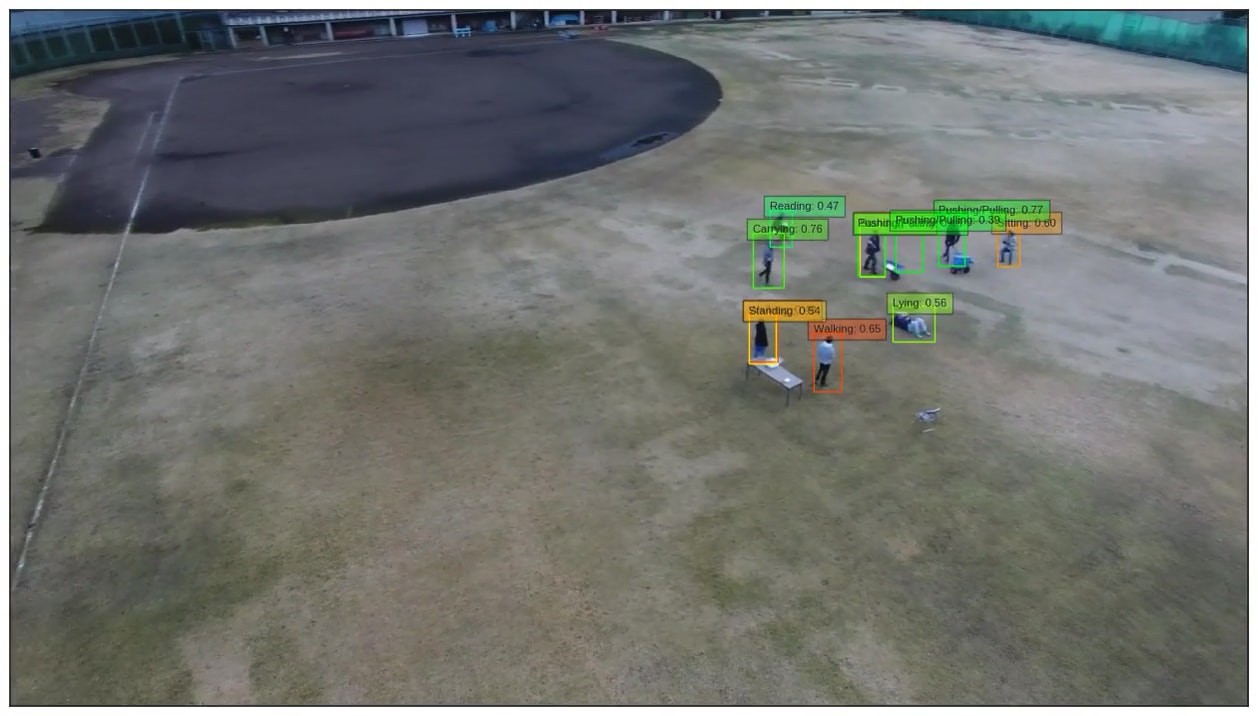}
     \caption{Sample detections of SSD 960x540 model for action detection. Detections with confidence score higher than 0.3 are shown. Each color corresponds to an action category. Frames from Okutama-Action test set.}
     \label{fig:actdetsample}
\end{figure*}

\textbf{Results} Table \ref{tab:actionresults} shows the mAP of action detection on the test set of Okutama-Action. As in object detection, mAP@0.5 is commonly used as an evaluation metric for action detection \cite{saha2016deep}. Figure \ref{fig:perclassres} shows the results for each class for the models trained on the natural RGB images, comparing both input sizes. An improvement with increased input size can be seen for most of the classes. Figure \ref{fig:actdetsample} shows selected detection results from the SSD 960x540 model for the action detection task. By visual inspection we observe that the performance is better when the camera angle is 45 degrees. The reason for this may be that with a lower angle, each actor occupies more pixels of the image. As pointed out in \cite{liu2016ssd}, our models are good at localizing objects, but worse at distinguishing classes. The gap between mAP for pedestrian detection and action detection confirms this. We see that the actions strongly related to temporal aspects have a low accuracy, e.g.\ \textit{Running} often being confused with \textit{Walking}. This is likely because we only distinguish classes at a frame-by-frame level. On the other hand, both \textit{Pushing} and \textit{Carrying} are more easily classified, which we believe is due to the fact that there is a large object next to the actor. Table \ref{tab:dataset_comparison} shows how the results we achieve on Okutama-Action compare to the best reported results on other datasets \cite{singh2016online}. 

\begin{table}[htb]
    \centering
    \begin{tabular}{|c|c|c|}
        \hline
        Image           &   Size    &   mAP ($\%$)      \\ \hline
        RGB             &   512x512 &   15.39           \\
        RGB             &   960x540 &   \textbf{18.80}  \\
        Optical Flow    &   512x512 &   6.47            \\ \hline
    \end{tabular}
    \caption{Results of different SSD models for action detection on Okutama-Action test set. The mAP is computed at 0.5 IoU threshold.}
    \label{tab:actionresults}
\end{table}

\begin{table}[htb]
    \centering
    \begin{tabular}{|c|c|c|}
        \hline
        Dataset         &   Model description   &   mAP ($\%$)  \\ \hline
        UCF101          &   SSD, 300x300, RGB \cite{saha2016deep}   &   37.93   \\
        J-HMDB          &   SSD, 300x300, RGB \cite{saha2016deep}   &   48.39   \\
        Okutama-Action  &   SSD, 960x540, RGB                       &   18.80   \\ \hline
    \end{tabular}
    \caption{Best reported results for different action detection datasets. The mAP is computed at 0.5 IoU threshold. }
    \label{tab:dataset_comparison}
\end{table}

\section{Conclusions}
\label{sec:conclusions}
In this paper, we present Okutama-Action, a new dataset for concurrent human action detection. It is a high resolution aerial view dataset consisting of 43 minute-long sequences, with 12 different action classes. By comparing performance in action detection with existing datasets, we show that Okutama-Action is more challenging, due to having non-static carema with abrupt motion, dynamic transition of actions, multiple concurrent actions and multi-labeled actors. An action detection model is trained and evaluated on our dataset, the performance of which demonstrates the difficulty of the dataset. In order to motivate more research in this area, we plan to make our dataset publicly available at \href{http://okutama-action.org}{\texttt{okutama-action.org}}.

As future work, we wish to adopt deep learning models that can handle multi-labeled outputs, to address the multiple-action annotation set provided for Okutama-Action. Another area worth investigating is to evaluate the performance of Multiple Object Tracking algorithms on our dataset, which should not be trivial considering the existing challenges.

\ifcvprfinal
\section*{Acknowledgement}
Research in this paper was supported by National Institute of Informatics, Tokyo. We would like to thank researchers at Prendinger and Inamura laboratory for their participation in our data collection. Thanks also to Mr. Eduardo Cordeiro who contributed to the annotation of the dataset and to Ms. Marcia Baptista and Mr. Sergi Caelles for their feedback. The work was partially supported by Prof. Matsuo’s Grant-in-Aid for Scientific Research on Innovative Areas on “Improvement of Predictability by Integrating Deep Learning with Symbol Processing”.
\fi

{\small
\bibliographystyle{ieee}
\bibliography{bibliography}

\begin{thebibliography}{10}\itemsep=-1pt

\bibitem{activitynet}
F.~Caba~Heilbron, V.~Escorcia, B.~Ghanem, and J.~Carlos~Niebles.
\newblock Activitynet: A large-scale video benchmark for human activity
  understanding.
\newblock In {\em Proceedings of the IEEE Conference on Computer Vision and
  Pattern Recognition}, pages 961--970, 2015.

\bibitem{survey2013}
J.~M. Chaquet, E.~J. Carmona, and A.~Fern{\'a}ndez-Caballero.
\newblock A survey of video datasets for human action and activity recognition.
\newblock {\em Computer Vision and Image Understanding}, 117(6):633--659, 2013.

\bibitem{Huang2016SpeedaccuracyTF}
J.~Huang, V.~Rathod, C.~Sun, M.~Zhu, A.~K. Balan, A.~Fathi, I.~Fischer,
  Z.~Wojna, Y.~Song, S.~Guadarrama, and K.~Murphy.
\newblock Speed/accuracy trade-offs for modern convolutional object detectors.
\newblock {\em CoRR}, abs/1611.10012, 2016.

\bibitem{Jhmdb}
H.~Jhuang, J.~Gall, S.~Zuffi, C.~Schmid, and M.~J. Black.
\newblock Towards understanding action recognition.
\newblock In {\em Proceedings of the IEEE International Conference on Computer
  Vision}, pages 3192--3199, 2013.

\bibitem{thumos13}
Y.~Jiang, J.~Liu, A.~R. Zamir, G.~Toderici, I.~Laptev, M.~Shah, and
  R.~Sukthankar.
\newblock Thumos challenge: Action recognition with a large number of classes,
  2014.

\bibitem{action-survey}
S.~M. Kang and R.~P. Wildes.
\newblock Review of action recognition and detection methods.
\newblock {\em arXiv preprint arXiv:1610.06906}, 2016.

\bibitem{Sports-1M}
A.~Karpathy, G.~Toderici, S.~Shetty, T.~Leung, R.~Sukthankar, and L.~Fei-Fei.
\newblock Large-scale video classification with convolutional neural networks.
\newblock In {\em Proceedings of the IEEE conference on Computer Vision and
  Pattern Recognition}, pages 1725--1732, 2014.

\bibitem{hmdb}
H.~Kuehne, H.~Jhuang, E.~Garrote, T.~Poggio, and T.~Serre.
\newblock Hmdb: a large video database for human motion recognition.
\newblock In {\em Computer Vision (ICCV), 2011 IEEE International Conference
  on}, pages 2556--2563. IEEE, 2011.

\bibitem{Hollywood1}
I.~Laptev, M.~Marszalek, C.~Schmid, and B.~Rozenfeld.
\newblock Learning realistic human actions from movies.
\newblock In {\em Computer Vision and Pattern Recognition, 2008. CVPR 2008.
  IEEE Conference on}, pages 1--8. IEEE, 2008.

\bibitem{liu2016ssd}
W.~Liu, D.~Anguelov, D.~Erhan, C.~Szegedy, S.~Reed, C.-Y. Fu, and A.~C. Berg.
\newblock Ssd: Single shot multibox detector.
\newblock In {\em European Conference on Computer Vision}, pages 21--37.
  Springer, 2016.

\bibitem{Hollywood2}
M.~Marszalek, I.~Laptev, and C.~Schmid.
\newblock Actions in context.
\newblock In {\em Computer Vision and Pattern Recognition, 2009. CVPR 2009.
  IEEE Conference on}, pages 2929--2936. IEEE, 2009.

\bibitem{Olympic-sports}
J.~C. Niebles, C.-W. Chen, and L.~Fei-Fei.
\newblock Modeling temporal structure of decomposable motion segments for
  activity classification.
\newblock In {\em European conference on computer vision}, pages 392--405.
  Springer, 2010.

\bibitem{virat}
S.~Oh, A.~Hoogs, A.~Perera, N.~Cuntoor, C.-C. Chen, J.~T. Lee, S.~Mukherjee,
  J.~Aggarwal, H.~Lee, L.~Davis, et~al.
\newblock A large-scale benchmark dataset for event recognition in surveillance
  video.
\newblock In {\em Computer vision and pattern recognition (CVPR), 2011 IEEE
  conference on}, pages 3153--3160. IEEE, 2011.

\bibitem{redmon2016you}
J.~Redmon, S.~Divvala, R.~Girshick, and A.~Farhadi.
\newblock You only look once: Unified, real-time object detection.
\newblock In {\em Proceedings of the IEEE Conference on Computer Vision and
  Pattern Recognition}, pages 779--788, 2016.

\bibitem{stanford}
A.~Robicquet, A.~Sadeghian, A.~Alahi, and S.~Savarese.
\newblock Learning social etiquette: Human trajectory understanding in crowded
  scenes.
\newblock In {\em European Conference on Computer Vision}, pages 549--565.
  Springer, 2016.

\bibitem{ucf-sports-1}
M.~D. Rodriguez, J.~Ahmed, and M.~Shah.
\newblock Action mach a spatio-temporal maximum average correlation height
  filter for action recognition.
\newblock In {\em Computer Vision and Pattern Recognition, 2008. CVPR 2008.
  IEEE Conference on}, pages 1--8. IEEE, 2008.

\bibitem{cooking-2}
M.~Rohrbach, S.~Amin, M.~Andriluka, and B.~Schiele.
\newblock A database for fine grained activity detection of cooking activities.
\newblock In {\em Computer Vision and Pattern Recognition (CVPR), 2012 IEEE
  Conference on}, pages 1194--1201. IEEE, 2012.

\bibitem{cooking-1}
M.~Rohrbach, A.~Rohrbach, M.~Regneri, S.~Amin, M.~Andriluka, M.~Pinkal, and
  B.~Schiele.
\newblock Recognizing fine-grained and composite activities using hand-centric
  features and script data.
\newblock {\em International Journal of Computer Vision}, 119(3):346--373,
  2016.

\bibitem{UT-in}
M.~S. Ryoo and J.~K. Aggarwal.
\newblock Spatio-temporal relationship match: Video structure comparison for
  recognition of complex human activities.
\newblock In {\em Computer vision, 2009 ieee 12th international conference on},
  pages 1593--1600. IEEE, 2009.

\bibitem{saha2016deep}
S.~Saha, G.~Singh, M.~Sapienza, P.~H. Torr, and F.~Cuzzolin.
\newblock Deep learning for detecting multiple space-time action tubes in
  videos.
\newblock {\em arXiv preprint arXiv:1608.01529}, 2016.

\bibitem{shu2015}
T.~Shu, D.~Xie, B.~Rothrock, S.~Todorovic, and S.~Chun~Zhu.
\newblock Joint inference of groups, events and human roles in aerial videos.
\newblock In {\em Proceedings of the IEEE Conference on Computer Vision and
  Pattern Recognition}, pages 4576--4584, 2015.

\bibitem{Simonyan14c}
K.~Simonyan and A.~Zisserman.
\newblock Very deep convolutional networks for large-scale image recognition.
\newblock {\em arXiv preprint arXiv:1409.1556}, 2014.

\bibitem{merl}
B.~Singh, T.~K. Marks, M.~Jones, O.~Tuzel, and M.~Shao.
\newblock A multi-stream bi-directional recurrent neural network for
  fine-grained action detection.
\newblock In {\em Proceedings of the IEEE Conference on Computer Vision and
  Pattern Recognition}, pages 1961--1970, 2016.

\bibitem{singh2016online}
G.~Singh, S.~Saha, and F.~Cuzzolin.
\newblock Online real time multiple spatiotemporal action localisation and
  prediction on a single platform.
\newblock {\em arXiv preprint arXiv:1611.08563}, 2016.

\bibitem{ucf-sports-2}
K.~Soomro and A.~R. Zamir.
\newblock Action recognition in realistic sports videos.
\newblock In {\em Computer Vision in Sports}, pages 181--208. Springer, 2014.

\bibitem{ucf101}
K.~Soomro, A.~R. Zamir, and M.~Shah.
\newblock Ucf101: A dataset of 101 human actions classes from videos in the
  wild.
\newblock {\em arXiv preprint arXiv:1212.0402}, 2012.

\bibitem{VATIC}
C.~Vondrick, D.~Patterson, and D.~Ramanan.
\newblock Efficiently scaling up crowdsourced video annotation.
\newblock {\em International Journal of Computer Vision}, 101(1):184--204,
  2013.

\bibitem{vu2014predicting}
T.-H. Vu, C.~Olsson, I.~Laptev, A.~Oliva, and J.~Sivic.
\newblock Predicting actions from static scenes.
\newblock In {\em European Conference on Computer Vision}, pages 421--436.
  Springer, 2014.

\bibitem{DALY}
P.~Weinzaepfel, X.~Martin, and C.~Schmid.
\newblock Towards weakly-supervised action localization.
\newblock {\em arXiv preprint arXiv:1605.05197}, 2016.

\bibitem{LIRIS}
C.~Wolf, E.~Lombardi, J.~Mille, O.~Celiktutan, M.~Jiu, E.~Dogan, G.~Eren,
  M.~Baccouche, E.~Dellandr{\'e}a, C.-E. Bichot, et~al.
\newblock Evaluation of video activity localizations integrating quality and
  quantity measurements.
\newblock {\em Computer Vision and Image Understanding}, 127:14--30, 2014.

\end{thebibliography}
}

\end{document}